\documentclass[10pt,twocolumn,letterpaper]{article}

\usepackage[pagenumbers]{latex/cvpr} 

\usepackage{graphicx}
\usepackage{amsmath}
\usepackage{amssymb}
\usepackage{booktabs}
\usepackage{algorithm}
\usepackage{algpseudocode}
\newcommand\blfootnote[1]{%
  \begingroup
  \renewcommand\thefootnote{}\footnote{#1}%
  \addtocounter{footnote}{-1}%
  \endgroup
}

\usepackage{amsmath}

\DeclareMathOperator*{\argmin}{arg\,min}

\usepackage[pagebackref,breaklinks,colorlinks]{hyperref}

\usepackage[capitalize]{cleveref}
\crefname{section}{Sec.}{Secs.}
\Crefname{section}{Section}{Sections}
\Crefname{table}{Table}{Tables}
\crefname{table}{Tab.}{Tabs.}

\def\cvprPaperID{52} 
\def\confName{CVPR}
\def\confYear{2022}

\begin{document}

\title{On Conditioning the Input Noise for Controlled Image Generation with Diffusion Models}

\author{Vedant Singh* \\
IIT Hyderabad\\
{\tt\small cs18btech11047@iith.ac.in}
\and
Surgan Jandial* \\
Adobe MDSR\\
{\tt\small jandialsurgan@gmail.com }
\and
Ayush Chopra\\
MIT\\
{\tt\small ayushc@mit.edu}
\and
Siddharth Ramesh\\
Adobe MDSR\\
{\tt\small sir@adobe.com}
\and
Balaji Krishnamurthy\\
Adobe MDSR\\
{\tt\small kbalaji@adobe.com}
\and
Vineeth N. Balasubramanian\\
IIT Hyderabad\\
{\tt\small vineethnb@cse.iith.ac.in}
}
\maketitle  

\begin{abstract}
Conditional image generation has paved the way for several breakthroughs in image editing, generating stock photos and 3-D object generation. This continues to be a significant area of interest with the rise of new state-of-the-art methods that are based on diffusion models. However, diffusion models provide very little control over the generated image, which led to subsequent works exploring techniques like classifier guidance, that provides a way to trade off diversity with fidelity. In this work, we explore techniques to condition diffusion models with carefully crafted input noise artifacts. This allows generation of images conditioned on semantic attributes. This is different from existing approaches that input Gaussian noise and further introduce conditioning at the diffusion model's inference step. Our experiments over several examples and conditional settings show the potential of our approach.
\end{abstract}

\blfootnote{* equal contribution}

\section{Introduction}
Deep Networks have resulted in significant breakthroughs over the recent times. Among the varied use cases, AI-powered creative content generation garners a lot of interest from a diverse set of stakeholders, right from artists to researchers. They have not only helped in speeding up the creation process but also in uncovering visual patterns beyond human imagination. The market for generative art is now worth millions of dollars which continues to grow with ongoing developments in NFTs and Metaverse. \\
Generative Adversarial Networks (GANs) have been at the forefront of numerous image generation tasks ever since their inception in the seminal work \cite{gan_goodfellow}. Over the past few years, they have outgrown their intended purpose of image synthesis and are now being used for specialized problems like image editing \cite{editgan, styleclip}, virtual try-on \cite{dior, sievenet, vitonsurvey}, or even making a pizza \cite{pizzagan}. Additionally, variants of GANs are able to explicitly control the class and features of the synthesized image. This, in formal parlance, is referred to as "Conditional Image Generation". Methods in literature have proposed several such approaches, for instance, \cite{cgan, acgan} explored conditional image generation using class as condition, for example, \cite{pix2pix} explored image-to-image translations, \cite{gatys_style} did style transfer, \cite{ledig} did image super resolution. \\
\begin{figure}
    \centering
    \includegraphics[width=0.9\linewidth]{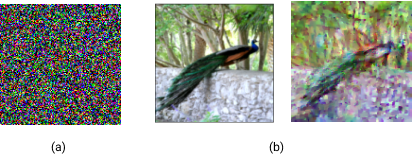}
    \caption{Comparison of (a) Random Gaussian Noise used in Diffusion Models and (b) Object Saliency Noise corresponding to the given reference image. We note (b) to highlight the object region, which supports our approach.}
    \label{fig:noise_comp}
\end{figure}
Alternatively, recent works have shown a far superior generation with diffusion models. They are a family of models which learn image generation by first progressively noising the image, slowly wiping out details till it becomes pure noise, and then training the model to iteratively reverse this noising process. To generate images after training, diffusion models start with a Gaussian noise and apply their iterative denoising process until a clean image is produced. Diffusion models have several important advantages over existing model families: GAN-level sample quality without adversarial training, analytic tractability and flexible model architecture.
Albeit the efficacy, the use of diffusion models for image synthesis is still fairly new, and unlike GANs, they don't offer much control over generation. In this work, we discuss controllability in the context of Diffusion Models. Recent works \cite{llvr} and \cite{morecontrol} propose the use of specific guidance functions on a pre-trained diffusion model to increase the control. More precisely, they take the output at each generation step, apply their guidance function to it and repeat this process iteratively through all the generation steps, thus imposing conditioning on the final output image by conditioning the output sample at each image generation step. Looking at it differently, we explore the problem of controlling the diffusion model output by only controlling its input noise. In other words, we intend to create a specialized noise which offers some control in itself and does not require further guidance at each generation step. Our initial experiments over multiple image classes and conditions reinforce our intuition that conditioning input noise can lead to better control over localization and saliency.\\
\label{sec:intro}

\section{Denoising Diffusion Probabilistic Models}
In this section, we provide a brief overview of DDPMs \cite{ddpm}, which we use for the purpose of our approach. On a high level, DDMPs work by progressively degrading the image $x$ for $T$ time steps with Gaussian noise and then training a neural network to reverse the gradual noising process. During sampling, the model synthesizes data from pure noise using the learnt denoisng process. In particular, sampling starts with noise $x_T$ and produces gradually less-noisy samples $x_{T-1}, x_{T-2}, \cdots$
until reaching a final sample $x_0$

The forward noising process $q$ at each iteration step is given as:
\begin{equation}
    q(x_{t} | x_{t-1}) = N(x_{t}; \sqrt{1 - \beta_{t}} x_{t-1}, \beta_{t} \textbf{I})
\end{equation}
where \textbf{I} is the Identity Matrix and $\beta_{t}$ is the constant defining the schedule of added noise. 

\begin{figure}[h]
    \centering
    \begin{tabular}{c}
            \includegraphics[width=\linewidth]{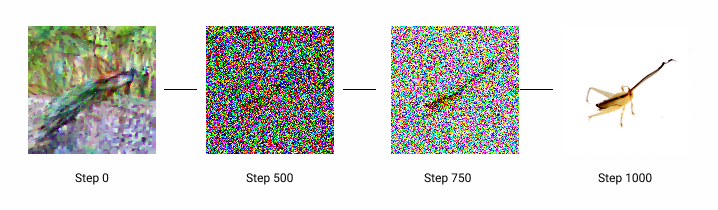} \\
        \vspace{-10pt}
        \includegraphics[width=\linewidth]{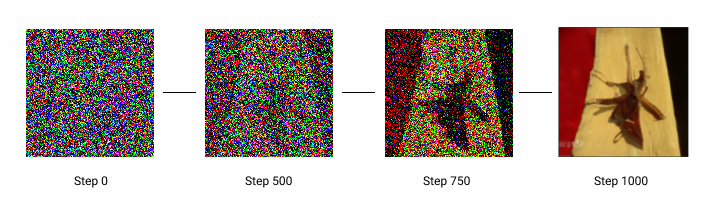} \\
    \end{tabular}
    \caption{Visualization of Diffusion Model generation with Random Noise (Row-1) and Ours (Row-2). Note that our noise includes salient regions than being completely random as Row-1.}
    \label{fig:noise_vis}
\end{figure}

\cite{ddpm} describes an $\alpha_{t} = 1 - \beta_{t}$, $\Bar{\alpha_{t}} = \prod_{i = 0}^{t} \alpha_{t}$, and then sampling at an arbitrary timestep $t$ as:
\begin{equation}
    q(x_{t} | x_{0}) = N(x_{t}; \sqrt{\Bar{\alpha_{t}}} x_{0}, (1 - \Bar{\alpha_{t}}) \textbf{I})
\end{equation}
which is then reparametrized as:
\begin{equation}
    x_t = \sqrt{\Bar{\alpha_{t}}} x_0 + (1 - \Bar{\alpha_{t}})\epsilon, \epsilon \sim N (0, I)
\end{equation}
The reverse process $p$, parameterized by a $\theta$ is thus given as:
\begin{equation}
    p_{\theta}(x_{t-1} | x_{t}) = N (x_{t-1}; \mu_{\theta}(x_{t}, t), {\sigma_{\theta}}^{2}(x_t, t)) 
\end{equation}
Further, it is shown in \cite{ddpm} that $x_{t-1}$ can be predicted from $x_{t}$ as:
\begin{equation}\label{eqn:denoising}
    x_{t-1} = \frac{1}{\sqrt{\alpha_{t}}} (x_{t} - \frac{1 - \alpha_{t}}{\sqrt{1-\Bar{\alpha_{t}}}}
\epsilon_{\theta} (x_{t}, t)) + \sigma_{\theta} z,  z \sim N (0, \textbf{I}) 
\end{equation}
Note that the parametrized $\epsilon_{\theta}$'s is the network learnt for denoising. \cite{praful_ddpm} uses a UNet \cite{unet} for the above. Fig. \ref{fig:noise_vis} (Row-2) visualizes the outputs over different diffusion model time-steps at inference.
\section{Guiding the Input Noise for Conditional Generation}
From Fig.\ref{fig:noise_comp}(a), we can visualize the default input noise that is passed through the DDPMs. In this section, we discuss our technique to condition the input noise and show how it leads to control over image generation in diffusion models. We begin by describing the characteristics of our noise.

\begin{figure*}
    \centering
    \includegraphics[scale=.45]{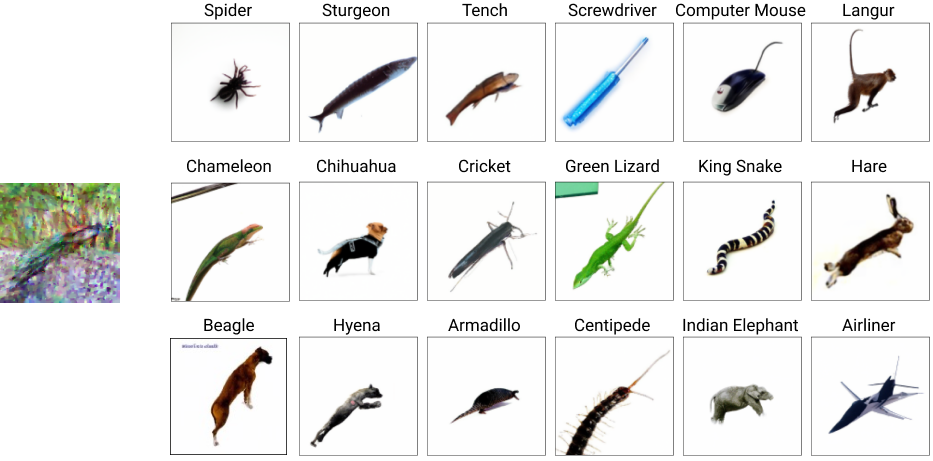}
    \caption{Results of Image Generation using Object Saliency Noise. We observe the Diffusion Model to output images in the constrained orientation for multiple classes.}
    \label{fig:main_results}
\end{figure*}

\subsection{Object Saliency Noise Guidance} 
In contrast to the completely random noise used in diffusion models, we propose the use of a noise which attends to salient regions representing the object. This noise can be visualised in Fig \ref{fig:noise_comp}(b). Note that the noise in this example is generated for the input image of a "peacock". We clearly see that this modified noise captures the saliency and orientation of regions where the peacock exists. \\
We show that when this noise is supplied to the diffusion model, it generates an image with the same localization and orientation as the noise, thus conditioning the output by conditioning the noise. \\
With the above motivation, we intend the generated noise to conform to the following characteristics: \\
\noindent \textbf{Salient Regions.} As discussed earlier, the noise should attend to the salient regions and guide the localization/orientation in the generated image. \\
\noindent \textbf{Same Range of Input Values.} As we don't re-train the diffusion model, we seek a noise which follows the same input space as comprehensible by the diffusion model. To be precise, diffusion model starts with a random Gaussian noise $\sim N(0, I)$ and updates using Eq. \ref{eqn:denoising} subsequently. 

\subsection{Inverting Gradients for Noise Generation}
We propose the use of Inverting Gradients (IG) \cite{ig} to obtain a noise with the above characteristics. \\
IG begins with a random Gaussian noise and iteratively updates it to re-generate the real image. The image-space update of the intermediate image $x_{T}$ at each step happens such that the gradient of  $x_{T}$ is similar to the gradient of the desired real image $I_{R}$. \\
More formally, let \(x^* \) be the original real image that needs to be recreated, \( y\) be its true class and \( \mathcal{L}_{\theta}(x^*, y) \) be the loss of the network's prediction for \( x^* \) under the parameters \( \theta \). For the sake of simplicity, the optimization problem for IG can be given as:
\begin{equation}
\label{eq:D}
\argmin_{x} {1 - \frac{\langle \nabla_{\theta} \mathcal{L}_{\theta}(x, y) , \nabla_{\theta} \mathcal{L}_{\theta}((x^*, y) \rangle}{\left \| \nabla_{\theta} \mathcal{L}_{\theta}((x, y) \right \| \left \| \nabla_{\theta} \mathcal{L}_{\theta}((x^*, y) \right \|}}
\end{equation} 
Thus, we find an image \( x \) whose gradients have the maximum possible cosine similarity with the gradients of our desired image \( x^* \). In other words, we successively update the intermediate $x_{T}$ according to the optimization problem mentioned in Eq \ref{eq:D}. Fig. \ref{fig:noise_influence} (Col-1) shows the progression of noise over steps. Note that, since we intend to use the input noise merely for localization, we choose the noisy samples from intermediate steps rather than the final generated image. We now discuss how the intermediate outputs of IG \cite{ig} suit our purpose:

\noindent\textbf{IG maintains Salient Regions.} IG \cite{ig} updates the noise by comparing the gradients of the generated image and the desired image. As gradients most likely account for the salient regions, we can inherently believe this comparison step largely involves updating noise around the salient regions. Hence, conforming to the first desired characteristic.

\noindent\textbf{IG has similar range of input values.} IG \cite{ig} begins with a Random Gaussian $\sim N(0, I)$ and updates it for generation, thus intuitively, this finds similarities to Eq. \ref{eqn:denoising}, wherein DDPMs \cite{ddpm, praful_ddpm} also updates the random noise based on an optimization. We hence hypothesize the intermediate IG outputs to be a comprehensible input for diffusion models. These claims find support in our results.

\begin{algorithm}[h]
\caption{Generation Using Object Saliency Noise}\label{alg:loc}
\begin{algorithmic}
\Require Classifier loss function \( \mathcal{L}_{\theta} \), IG model \( \mathcal{I}\), inverting steps \( k \), diffusion model $ \mathcal{D}_{\phi} $, localisation guiding image $ x^* $, ground truth class \( y \), target class \( t \) \\

\State \( grad \gets \nabla_{\theta} \mathcal{L}_{\theta} (x^*, y) \) 
\State \( x \gets \mathcal{I} (\mathcal{L}_{\theta}, grad, y, k) \) 

\State \( x' \gets \frac{x - \mu}{\sigma} \) 

\State \( out \gets \mathcal{D}_{\phi} (x', t) \) \\
\Return \( out \)

\end{algorithmic}
\end{algorithm}




\section{Experiments and Analysis}
\begin{figure}
    \centering
    \includegraphics[width=\linewidth]{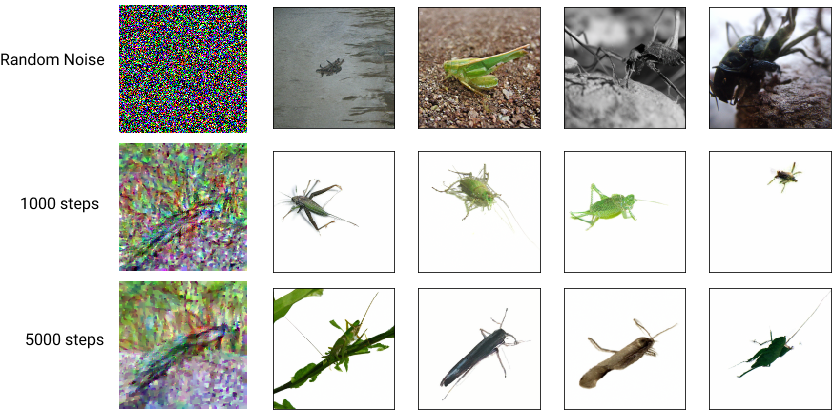}
    \caption{Effect of sampling outputs over different steps of IG. Col-1 visualizes IG outputs over steps- 0, 1000, 5000 steps. We clearly see the initial IG output- 0 step lacks localization, whereas the localisation provided by 1000 step is not as sound as in 5000 step.}
    \label{fig:noise_influence}
\end{figure}

We begin this section by defining the baseline models and later present our results. 
For the initial noise modification, we use the Inverting Gradients \cite{ig} method that has been trained on ImageNet \cite{imagenet} with ResNet18 architecture. We run this method for \( k \in \{1000, 3000, 5000 \} \) steps and further show its influence on generation in Fig. \ref{fig:noise_influence}. In the next step of image generation, we use a pre-trained class-conditioned DDPM model \cite{praful_ddpm} that generates images of size 128x128 and has also been trained on ImageNet.


\subsection{Results}
This section includes our image generation results. We broadly demonstrate two instances of discussions. \\
\noindent\textbf{Generation across multiple classes.} Fig. \ref{fig:main_results} shows the generation results over multiple classes and with different samples of input noise (see (a), (b)).  We can clearly see that using the same input noise we can generate orientation/saliency conditioned images across multiple classes.\\
\noindent\textbf{Generation across noise manipulations.} In this experiment, we manipulate the noise with rotation, flipping and observe the results. We can see the changes in noise structurally influence the changes in diffusion model outputs.  \\
\begin{figure}
    \centering
    \includegraphics[width=\linewidth]{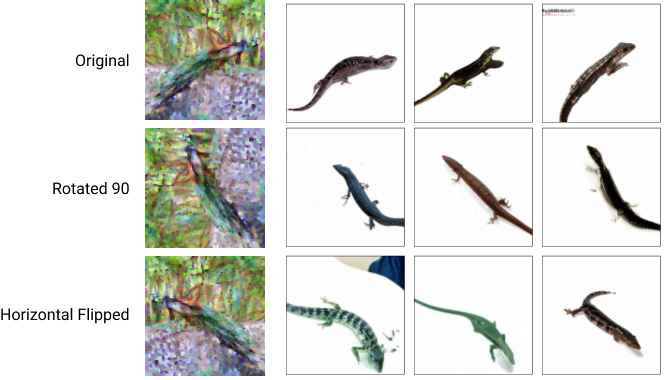}
    \caption{Outputs after manipulating input noise: (a) 90 degree rotation, (b) Horizontal flipping. Note the orientation of generated images change accordingly with the rotation, flip.}
    \label{fig:img_rotation}
\end{figure}

\subsection{Sampling from different steps of IG} Fig. \ref{fig:noise_influence} (Col-1) shows the intermediate step images generated in Inverting Gradients. As we discussed, our approach needs a noise that attends well to the object while maintaining the orientation. Therefore, in Fig. \ref{fig:noise_influence}, we also see that the generation significantly improves for 5000 steps over 1000 steps. This is primarily because the noise of 1000 steps has a poor localization in comparison to noise of 5000 steps.

\subsection{Conclusion and Future Work}

\begin{figure}
    \centering
    \includegraphics[width=\linewidth]{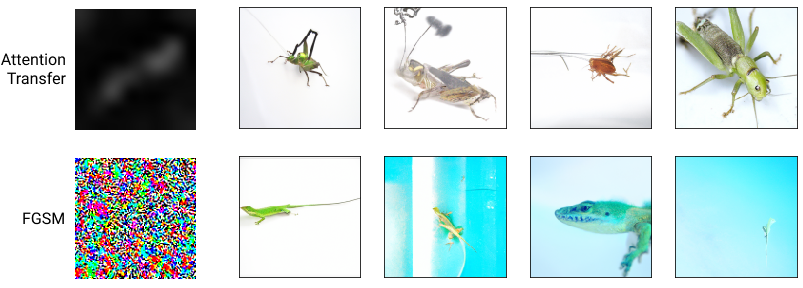}
    \caption{Generation Results using inputs from 1) ResNet Feature Maps 2) FGSM Gradient Maps. We can clearly see the above images are not as well defined in \ref{fig:main_results}.}
    \label{fig:other_sal}
\end{figure}
In this work, we explore the idea of controlling diffusion model output by creating a specialized form of noise, which we call Object Saliency Noise. Our results support this proposition and as part of our future work, we plan the following: 
\begin{itemize}
\item To benefit from the localization, we can also use several other ways in current literature such as \cite{at, fgsm} to obtain the diffusion model inputs. Thus, as part of our first efforts in this direction, we tried using Saliency maps obtained by channel averaging ResNet Features Maps as in \cite{at} and Gradient Maps obtained by FGSM \cite{fgsm}. Fig \ref{fig:other_sal} includes these results. We note that the one difference between these methods and IG might be that the former may not have similar range of input values as diffusion models because of their definition. We plan to study this exhaustively.
\item One key observation from the results in the paper is that the background of the image is often inhibited. Thus, as our future work, we plan to study this nuance of Object Saliency Noise to potentially generate background free noise.

\end{itemize}

{\small
\bibliographystyle{ieee_fullname}
\bibliography{egbib}

\begin{thebibliography}{10}\itemsep=-1pt

\bibitem{llvr}
Jooyoung Choi, Sungwon Kim, Yonghyun Jeong, Youngjune Gwon, and Sungroh Yoon.
\newblock Ilvr: Conditioning method for denoising diffusion probabilistic
  models.
\newblock In {\em Proceedings of the IEEE/CVF International Conference on
  Computer Vision (ICCV)}, pages 14367--14376, October 2021.

\bibitem{dior}
Aiyu Cui, Daniel McKee, and Svetlana Lazebnik.
\newblock Dressing in order: Recurrent person image generation for pose
  transfer, virtual try-on and outfit editing.
\newblock In {\em Proceedings of the IEEE/CVF International Conference on
  Computer Vision (ICCV)}, pages 14638--14647, October 2021.

\bibitem{imagenet}
Jia Deng, Wei Dong, Richard Socher, Li-Jia Li, Kai Li, and Li Fei-Fei.
\newblock Imagenet: A large-scale hierarchical image database.
\newblock In {\em 2009 IEEE Conference on Computer Vision and Pattern
  Recognition}, pages 248--255, 2009.

\bibitem{praful_ddpm}
Prafulla Dhariwal and Alexander~Quinn Nichol.
\newblock Diffusion models beat {GAN}s on image synthesis.
\newblock In A. Beygelzimer, Y. Dauphin, P. Liang, and J.~Wortman Vaughan,
  editors, {\em Advances in Neural Information Processing Systems}, 2021.

\bibitem{gatys_style}
Leon~A. Gatys, Alexander~S. Ecker, and Matthias Bethge.
\newblock Image style transfer using convolutional neural networks.
\newblock In {\em 2016 IEEE Conference on Computer Vision and Pattern
  Recognition (CVPR)}, pages 2414--2423, 2016.

\bibitem{ig}
Jonas Geiping, Hartmut Bauermeister, Hannah Dr\"{o}ge, and Michael Moeller.
\newblock Inverting gradients - how easy is it to break privacy in federated
  learning?
\newblock In H. Larochelle, M. Ranzato, R. Hadsell, M.~F. Balcan, and H. Lin,
  editors, {\em Advances in Neural Information Processing Systems}, volume~33,
  pages 16937--16947. Curran Associates, Inc., 2020.

\bibitem{vitonsurvey}
Hajer Ghodhbani, Adel Alimi, Mohamed Neji, and Imran Razzak.
\newblock You can try without visiting: A comprehensive survey on virtually
  try-on outfits, Feb 2021.

\bibitem{gan_goodfellow}
Ian Goodfellow, Jean Pouget-Abadie, Mehdi Mirza, Bing Xu, David Warde-Farley,
  Sherjil Ozair, Aaron Courville, and Yoshua Bengio.
\newblock Generative adversarial nets.
\newblock In Z. Ghahramani, M. Welling, C. Cortes, N. Lawrence, and K.~Q.
  Weinberger, editors, {\em Advances in Neural Information Processing Systems},
  volume~27. Curran Associates, Inc., 2014.

\bibitem{fgsm}
Ian Goodfellow, Jonathon Shlens, and Christian Szegedy.
\newblock Explaining and harnessing adversarial examples.
\newblock In {\em International Conference on Learning Representations}, 2015.

\bibitem{pizzagan}
Fangda Han, Guoyao Hao, Ricardo Guerrero, and Vladimir Pavlovic.
\newblock Mpg: A multi-ingredient pizza image generator with conditional
  stylegans, 2020.

\bibitem{ddpm}
Jonathan Ho, Ajay Jain, and Pieter Abbeel.
\newblock Denoising diffusion probabilistic models, 2020.

\bibitem{pix2pix}
Phillip Isola, Jun-Yan Zhu, Tinghui Zhou, and Alexei~A Efros.
\newblock Image-to-image translation with conditional adversarial networks.
\newblock {\em CVPR}, 2017.

\bibitem{sievenet}
Surgan Jandial, Ayush Chopra, Kumar Ayush, Mayur Hemani, Balaji Krishnamurthy,
  and Abhijeet Halwai.
\newblock Sievenet: A unified framework for robust image-based virtual try-on.
\newblock In {\em Proceedings of the IEEE/CVF Winter Conference on Applications
  of Computer Vision (WACV)}, March 2020.

\bibitem{ledig}
Justin Johnson, Alexandre Alahi, and Li Fei-Fei.
\newblock Perceptual losses for real-time style transfer and super-resolution.
\newblock In Bastian Leibe, Jiri Matas, Nicu Sebe, and Max Welling, editors,
  {\em Computer Vision -- ECCV 2016}, pages 694--711, Cham, 2016. Springer
  International Publishing.

\bibitem{editgan}
Huan Ling, Karsten Kreis, Daiqing Li, Seung~Wook Kim, Antonio Torralba, and
  Sanja Fidler.
\newblock Editgan: High-precision semantic image editing.
\newblock In {\em Advances in Neural Information Processing Systems (NeurIPS)},
  2021.

\bibitem{morecontrol}
Xihui Liu, Dong~Huk Park, Samaneh Azadi, Gong Zhang, Arman Chopikyan, Yuxiao
  Hu, Humphrey Shi, Anna Rohrbach, and Trevor Darrell.
\newblock More control for free! image synthesis with semantic diffusion
  guidance.
\newblock 2021.

\bibitem{cgan}
Mehdi Mirza and Simon Osindero.
\newblock Conditional generative adversarial nets, 2014.
\newblock cite arxiv:1411.1784.

\bibitem{acgan}
Augustus Odena, Christopher Olah, and Jonathon Shlens.
\newblock Conditional image synthesis with auxiliary classifier gans.
\newblock In {\em Proceedings of the 34th International Conference on Machine
  Learning - Volume 70}, ICML'17, page 2642–2651. JMLR.org, 2017.

\bibitem{styleclip}
Or Patashnik, Zongze Wu, Eli Shechtman, Daniel Cohen-Or, and Dani Lischinski.
\newblock Styleclip: Text-driven manipulation of stylegan imagery.
\newblock In {\em Proceedings of the IEEE/CVF International Conference on
  Computer Vision (ICCV)}, pages 2085--2094, October 2021.

\bibitem{unet}
Olaf Ronneberger, Philipp Fischer, and Thomas Brox.
\newblock U-net: Convolutional networks for biomedical image segmentation.
\newblock In Nassir Navab, Joachim Hornegger, William~M. Wells, and
  Alejandro~F. Frangi, editors, {\em Medical Image Computing and
  Computer-Assisted Intervention -- MICCAI 2015}, pages 234--241, Cham, 2015.
  Springer International Publishing.

\bibitem{at}
Sergey Zagoruyko and Nikos Komodakis.
\newblock Paying more attention to attention: Improving the performance of
  convolutional neural networks via attention transfer.
\newblock In {\em ICLR}, 2017.

\end{thebibliography}
}

\end{document}